\documentclass{article}
\usepackage{ijcai16}
\usepackage{times}
\usepackage{color}
\usepackage{subfigure}
\usepackage{subfig}
\usepackage{graphicx}
\usepackage{bm}
\usepackage{subfig}
\usepackage{amsmath,amssymb}
\usepackage{graphics}
\usepackage{algorithm}
\usepackage{algorithmic}
\usepackage{url}
\usepackage{balance}
\usepackage{amsthm}
\usepackage{booktabs}

\newtheorem{theorem}{Theorem}
\title{$L_0$-norm Sparse Graph-regularized SVD for Biclustering}
\author{Wenwen Min$^1$, Juan Liu$^{1*}$, and Shihua Zhang$^{2*}$\\
                  $^1$School of Computer, Wuhan University, Wuhan 430072, China \\
                  $^2$National Center for Mathematics and Interdisciplinary Sciences, Academy of Mathematics\\
                  and Systems Science, Chinese Academy of Sciences, Beijing 100190, China\\
                  mww@whu.edu.cn, liujuan@whu.edu.cn and zsh@amss.ac.cn}	
\begin{document}
\maketitle
\begin{abstract}
Learning the ``blocking'' structure is a central challenge for high dimensional data (e.g., gene expression data). In \cite{lee2010biclustering}, a sparse singular value decomposition (SVD) has been used as a biclustering tool to achieve this goal. However, this model ignores the structural information between variables (e.g., gene interaction graph). Although typical graph-regularized norm can incorporate such prior graph information to get accurate discovery and better interpretability, it fails to consider the opposite effect of variables with different signs. Motivated by the development of sparse coding and graph-regularized norm, we propose a novel sparse graph-regularized SVD as a powerful biclustering tool for analyzing high-dimensional data. The key of this method is to impose two penalties including a novel graph-regularized norm ($|\bm{u}|\bm{L}|\bm{u}|$) and $L_0$-norm ($\|\bm{u}\|_0$) on singular vectors to induce structural sparsity and enhance interpretability. We design an efficient Alternating Iterative Sparse Projection (AISP) algorithm to solve it. Finally, we apply our method and related ones to simulated and real data to show its efficiency in capturing natural blocking structures.
\end{abstract}
\section{Introduction}
Singular Value Decomposition (SVD) is a powerful matrix decomposition model which has been widely used in different fields \cite{alter2000singular,aharon2006img,zhou2015svd}. Suppose $\bm{X}$ is a matrix of size $n \times p$ (e.g., a gene expression matrix), then the classical SVD of $\bm{X}$ is:
\begin{equation}\label{eq01}
   \bm{X} = \bm{UDV}^T = \sum_{k=1}^r {d_k \bm{u}_k\bm{v}_k},
\end{equation}
where $r$ is the rank of $\bm{X}$, $\bm{u_k}$ and $\bm{v_k} (k =1,\cdots, r$) are left and right singular vectors respectively, and both $\bm{U} = (\bm{u}_1,\cdots,\bm{u}_r) $ and $\bm{V} = (\bm{v}_1,\cdots,\bm{v}_r) $ are orthogonal matrices. The $\bm{D} = \mbox{diag}(d_1,d_2,\cdot\cdot\cdot,d_r)$ is a diagonal matrix. However, the non-sparse singular vectors ($\bm{u}_k$, $\bm{v}_k$) can sometimes be difficult to interpret. Many studies impose sparsity on singular vectors to improve this \cite{shi2013scmf,jing2015sparse}. The advantage of sparsity is that the sparse singular vectors are better able to capture inherent structures and patterns of the input data.

We here introduce a formal framework of the rank one sparse SVD \cite{witten2009penalized},
\begin{equation}
\begin{aligned}\label{equ:03}
& \underset{\bm{u},\bm{v},d}{\text{minimize}} && \|\bm{X} - d\bm{uv}^T\|_F^2\\
& \text{subject to}                  && \|\bm{u}\|_2 \leq 1, P_1(\bm{u}) \leq c_1,\\
&                                    && \|\bm{v}\|_2 \leq 1, P_2(\bm{v}) \leq c_2,
\end{aligned}
\end{equation}
where $P_1(\bm{u})$ and $P_2(\bm{v})$ are two penalty functions. For different purpose, different sparsity penalties can be induced in this framework \cite{witten2009penalized,lee2010biclustering,sill2011robust}. For example, Witten \emph{et~al.} \shortcite{witten2009penalized} has proposed a penalized matrix decomposition (PMD) by using $L_1$-penalty to induce sparsity \cite{witten2009penalized}. In addition, Lee \emph{et~al.} \shortcite{lee2010biclustering} has developed a sparse SVD (ALSVD) by imposing Adaptive Lasso penalty to the left/right singular vectors and employed it as a biclustering tool to capture bi-cluster structure of a gene expression data. However, these sparse SVD models have several limitations. First, they impose sparseness via $L_1$-norm penalty, while little work is concerned with $L_0$-norm one. Compared to $L_1$-norm, $L_0$-norm can enforce to get a desirable level of sparsity. Second, they ignore the prior graph knowledge for variables. Recent studies have shown that graph-regularized penalty can help to get better interpretability and higher generalization performance for regression tasks and others \cite{ando2007learning,li2009relation,li2008network,li2010variable,jiang2013graph}. To our knowledge, there is yet no study to incorporate graph structure into the sparse SVD framework.

In this paper, we propose a novel sparse graph-regularized SVD to model the high-dimensional data with prior graph knowledge. This method is inspired by recent developments of sparse coding and graph-regularized norm \cite{lee2006efficient,ando2007learning,zheng2011graph,he2011nonnegative}. It imposes a novel graph-regularized norm ($|\bm{u}|\bm{L}|\bm{u}|$) instead of the general one ($\bm{u}\bm{L}\bm{u}$) and the $L_0$-norm ($\|\bm{u}\|_0$) penalty on singular vectors to induce structural sparsity. In summary, our key contributions are threefold: (1) impose $L_0$-norm penalty to graph-regularized SVD (sgSVD); (2) consider smoothing of absolute signal in graph-regularized penalty (sgSVD*); (3) design an efficient Alternating Iterative Sparse Projection (AISP) algorithm to solve it. Finally, we apply sgSVD* and related methods to simulated and real data to demonstrate its efficiency in capturing blocking data structure.
\section{Related Works}
Recent studies have a growing interest in regularized SVD including the following main three aspects.

(1) Sparse SVD. Witten \emph{et~al.} \shortcite{witten2009penalized} proposed the first sparse SVD via $L_1$-norm and fused lasso as sparsity-inducing norm. Lee \emph{et~al.} \shortcite{lee2010biclustering} recommended to use different penalties, such as Lasso, adaptive Lasso and Smoothly Clipped Absolute Deviation (SCAD), as penalty functions. A natural choice of the penalty function is $L_0$ norm. Min \emph{et~al.} \shortcite{min2015novel} proposed a sparse SVD via $L_0$-norm penalty (L0SVD).

(2) Graph-regularized SVD. Graph-regularized norm has been used in many different techniques such as nonnegative matrix factorization \cite{cai2011graph}. However, to our knowledge, there is yet no study to incorporate graph-regularized norm into the sparse SVD framework. We believe that sparse graph-regularized SVD is a promising tool as explored in other problems \cite{li2009relation,jenatton2009structured,zheng2011graph} to enforce the smoothness of variable coefficients. In addition, $L_{2,1}$ is widely used to enforce the structural sparsity at the inter-group level \cite{jacob2009group}. We can also consider it as a penalty function of SVD in future studies.

(3) The relationship between SVD and PCA. As we all know, principal component analysis (PCA) can be efficiently solved by using SVD. However, the identified non-sparse principal components can sometimes be difficult to interpret. To solve it, recent studies have developed several different sparse PCA models \cite{deshpande2014sparse,gu2014sparse,shen2008sparse,witten2009penalized}. However, it is difficult to develop effective algorithms for solving these sparse PCA models. There are a kind of commonly used methods based on regularized SVD for solving them \cite{shen2008sparse}. Moreover, Witten \emph{et~al.} \shortcite{witten2009penalized} proposed another model based on regularized SVD to ensure the orthogonally of their left singular vectors.

\section{The Proposed Approach}
\subsection{Sparse Graph-regularized Penalty}
Given a simple graph $G$, its Laplacian matrix $\bm{L}$ is defined as $\bm{L} = \bm{D} - \bm{A}$, where $\bm{A}$ is the adjacency matrix of graph $G$, $\bm{A}_{ij}=1$ if vertex $i$ and $j$ is connected in the graph, $\bm{A}_{ij}=0$ otherwise, and $\bm{D}$ is a diagonal matrix whose diagonal elements $d_i$ is the degree of vertex $i$ (i.e., $d_i = \sum_{j=1}^p \bm{A}_{ij}$). We first consider the ordinary graph-regularized norm and impose sparsity on singular vectors in SVD framework with the following penalty,
\begin{equation}\label{equ:10}
  P(\bm{v}) = \lambda_1 \|\bm{v}\|_1+ \lambda_2 \bm{v}^T\bm{L}\bm{v},
\end{equation}
where $\lambda_1\geq 0$ and $\lambda_2 \geq 0$ are two regularization parameters. The first term $\|\bm{v}\|_1 = \sum_{i} |v_i|$ is the $L_1$-norm penalty to induce sparsity. The second term $\bm{v}^T\bm{L}\bm{v} = \frac{1}{2}\sum_{i\sim j}\bm{A}_{ij}(v_i-v_j)^2$ is a quadratic Laplacian penalty which forces the coefficients of $\bm{v}$ to be smooth. Thus, we can create a sparse graph-regularized SVD using the penalty Eq. (\ref{equ:10}) in the framework Eq. (\ref{equ:03}). The $L_1$-norm has some good nature and has been applied to induce sparsity in SVD \cite{witten2009penalized}. Whereas little work has been done using a more natural sparseness regularization, i.e., $L_0$-norm penalty. In this paper, we adopt $L_0$-norm penalty rather than $L_1$-norm one to induce sparsity on singular vectors. Thus, we replace Eq. (\ref{equ:10}) with the following one,
\begin{equation}\label{equ:SGR}
  P(\bm{v}) = \lambda_0 \|\bm{v}\|_0+ \lambda_2 \bm{v}^T\bm{L}\bm{v},
\end{equation}
where $\|\bm{v}\|_0 = \sum_{i} I(v_i)$ is the $L_0$-norm penalty.

The classical graph-regularized penalty in Eq. (\ref{equ:SGR}) ensures that two positively correlated variables would tend to influence their coefficients in the same direction. However, it makes two negatively correlated variables have opposite effect. Thus, the classical graph-regularized penalty may not perform well when the coefficients of a number of variables linked on the prior graph have different signs. This would lead to the absolute values of their coefficients are not expected to be locally smooth. To solve this drawback, we redefine a sparse graph-regularized penalty by only considering the magnitude of their coefficients,
\begin{equation}\label{equ:11}
  P(\bm{v}) = \lambda_1 \|\bm{v}\|_0+ \lambda_2 |\bm{v}|^T\bm{L}|\bm{v}|,
\end{equation}
where $|\bm{v}|^T\bm{L}|\bm{v}| = \frac{1}{2}\sum_{i\sim j}\bm{A}_{ij}(|v_i|-|v_j|)^2$ and $|\bm{v}|= (|v_1|,|v_2|,\cdots,|v_p|)^T$. This is also reflected by the fact that linked genes in a gene interaction graph are usually significantly positively or negatively correlated. This new penalty Eq. (\ref{equ:11}) encourages highly inter-correlated variables corresponding to a dense subnetwork in $G$ to be jointly selected.

Mathematically, the normalized Laplacian matrix $\widetilde{\bm{L}}$ defined by $\widetilde{\bm{L}} = \bm{D}^{-1/2}\bm{L}\bm{D}^{-1/2}$ can be used to replace $\bm{L}$ in Eq. (\ref{equ:10}), Eq. (\ref{equ:SGR}), and Eq. (\ref{equ:11}). The choice of $\bm{L}$ and normalized $\widetilde{\bm{L}}$ depends on the specific problems. In addition, there are a wide variety of graph matrix representations \cite{wilson2008study}, which can also be considered to replace $\bm{L}$. In this paper, we apply the sparse graph-regularized penalty with $\bm{L}$ for biclustering gene expression data.

\subsection{$L_1$-norm Sparse Graph-regularized SVD}
The sparse graph-regularized penalty via $L_0$-norm is non-convex, which leads to a challenge issue for efficient implementation. Here, we first discuss a sparse graph-regularized SVD by replacing $L_0$-norm in Eq.(\ref{equ:11}) with $L_1$-norm, denoted as $L_1$-sgSVD*. Formally, we should deal with the following minimization problem,
\begin{equation}
\begin{aligned}\label{equ:sgSVD*}
& \underset{\bm{u},\bm{v},d}{\text{minimize}} && \|\bm{X} - d\bm{uv}^T\|_F^2\\
& \text{subject to}    && \|\bm{u}\|_2 \leq 1, \|\bm{u}\|_1 \leq s_1, |\bm{u}|^T\bm{L_1}|\bm{u}| \leq s_2,\\
&     && \|\bm{v}\|_2 \leq 1, \|\bm{v}\|_1 \leq c_1, |\bm{v}|^T\bm{L_2}|\bm{v}| \leq c_2,
\end{aligned}
\end{equation}
where $d$ is a positive singular value, $\bm{u}$ is a $n \times 1$ column vector, $\bm{v}$ is a $p \times 1$ column vector, both $\bm{L_1}$ and $\bm{L_2}$ are Laplace matrices, and $s_1$, $s_2$, $c_1$ and $c_2$ are four super-parameters.

The objective function $\|\bm{X} - d\bm{uv}^T\|_F^2$ in Eq. (\ref{equ:sgSVD*}) is biconvex to $\bm{u}$ and $\bm{v}$. Thus, we propose an alternating iterative learning strategy to solve the optimization problem Eq. (\ref{equ:sgSVD*}). Next we consider optimization over $\bm{v}$ for a fixed $\bm{u}$, and get the following constrained optimization problem,
\begin{equation}
\begin{aligned}\label{equ:12}
& \underset{\bm{v},d}{\text{minimize}} && \|\bm{X} - d\bm{uv}^T\|_F^2\\
& \text{subject to}                  && \|\bm{v}\|_2 \leq 1,\\
&                                && \|\bm{v}\|_1 \leq c_1,\\
&                                && \bm{|v|}^T\bm{L}\bm{|v|} \leq c_2.
\end{aligned}
\end{equation}
We note that this objective function $||\bm{X}-d\bm{u}\bm{v}^t||_F^2 = \|\bm{X}\|_F^2-2d \bm{u}^T\bm{X}\bm{v}+d^2$. Obviously, minimizing this function $||\bm{X}-d\bm{u}\bm{v}^T||_F^2$ is equivalent to minimizing $-\bm{u}\bm{X}\bm{v}$. Let $\bm{z} = \bm{X}^T\bm{u}$. The problem Eq. (\ref{equ:12}) can thus be written as
\begin{equation}
\begin{aligned}\label{equ:13}
& \underset{\bm{v}}{\text{minimize}} && -\bm{v}^T\bm{z}\\
& \text{subject to}                  && \|\bm{v}\|_2 \leq 1,\\
&                                    && \|\bm{v}\|_1 \leq c_1,\\
&                                    && \bm{|v|}^T\bm{L}\bm{|v|} \leq c_2.
\end{aligned}
\end{equation}
The $\bm{|v|}^T\bm{L}\bm{|v|}$ is general non-convex, which leads to a challenge issue for solving the problem Eq. (\ref{equ:13}). Next we propose Theorem 1 to cleverly remove the absolute operator in the constrained condition.
\begin{theorem} Suppose $\bm{v}^*$ is an optimal solution of Eq.~(\ref{equ:13}), then $v^*_i z_i\geq 0$ for all $i$. \end{theorem}
\begin{proof}
Suppose the Theorem is false. Then there is at least exists $i$ to meet $v^*_i z_i < 0$. We first construct a vector $\bm{\widetilde{v}}$ which satisfies $\widetilde{v}_j = v^*_j$ for all $j \neq i$ and $\widetilde{v}_i = -v^*_i$. Obviously, the vector $\bm{\widetilde{v}}$ meets $\|\bm{\widetilde{v}}\|_0 = \|\bm{v}^*\|_0$ and $\|\bm{\widetilde{v}}\|_2 = \|\bm{v}^*\|_2$.
Let $f_1 = - \bm{v}^*\bm{z}$ and $f_2 = - \bm{\widetilde{v}}\bm{z}$, then $f_1 -f_2 = -2v^*_i z_i>0$ (i.e., $f_2 < f_1$). Thus, we get a vector $\bm{\widetilde{v}}$ which corresponds to a more smaller objective value than that of the optimal solution $\bm{v}^*$. It leads to a desired contradiction. Thus, the Theorem is true.
\end{proof}
Based on Theorem 1, we can replace Eq. (\ref{equ:13}) with the following optimization problem,
\begin{equation}
\begin{aligned}\label{equ:14}
& \underset{\bm{v}}{\text{minimize}} && -\bm{v}^T|\bm{z}|\\
& \text{subject to}                  && \|\bm{v}\|_2 \leq 1,\\
&                                    && \|\bm{v}\|_1 \leq c_1,\\
&                                    && \bm{v}^T\bm{L}\bm{v} \leq c_2,\\
&                                    && v_k \geq 0, \forall k,
\end{aligned}
\end{equation}
where $|\bm{z}|=(|z_1|,...,|z_p|)^T$. A standard way to solve the optimization problem Eq. (\ref{equ:14}) is to formulate the Lagrangian form as follows:
\begin{equation}\label{equ:15}
  \mathcal{L}(\bm{v}) = -\bm{v}^T|\bm{z}| + \lambda \sum_{i=1}^p v_i+ \frac{1}{2}\eta \bm{v}^T\bm{v}+\frac{1}{2} \sigma \bm{v}^T\bm{L}\bm{v}-\sum_{i=1}^p \tau_i v_i,
\end{equation}
where $\lambda \geq 0 , \eta \geq 0, \sigma \geq 0, \tau_i \geq 0$ are Lagrangian multipliers. Actually, we can easily get Theorem 2 to confirm the function Eq. (\ref{equ:15}) is a convex function (we omit the proof here).
\begin{theorem} The function Eq. (\ref{equ:15}) is a convex function. \end{theorem}
Thus, the optimal solution of Eq. (\ref{equ:15}) is characterized by its subgradient equation as follows:
\begin{equation}\label{equ:16}
  \nabla_{\bm{v}}\mathcal{L} = -|\bm{z}| + \lambda \bm{e} + \eta\bm{v}+\sigma \bm{L}\bm{v} - \bm{\tau} = \bm{0}.
\end{equation}
Ideally, we can get the optimal solution of Eq. (\ref{equ:15}) by solving the corresponding linear system in Eq. (\ref{equ:16}). However, this is inefficient due to the numerical difficulty of inverting a Hessian matrix. As an alternative way, we adopt a simple coordinate descent method \cite{friedman2007pathwise} to learn vector $\bm{v}$. The subgradient of $v_k$ in Eq. (\ref{equ:15}) is
\begin{equation}\label{equ:17}
  \frac{\partial \mathcal{L}}{\partial v_k} = -|z_k| + \lambda  + \eta v_k +\sigma d_k  v_k -\sigma  \bm{A}_k\bm{v}- \tau_k,
\end{equation}
where $d_k$ is the degree of node $k$ and $\bm{A}_k$ is the $k$th column of the adjacency matrix $\bm{A}$. The complementary slackness KKT condition implies that if $v_k > 0$, then $\tau_k = 0$ and if $v_k = 0$, then $\tau_k > 0$. Thus, let Eq. (\ref{equ:17}) be zero, then
\begin{equation}\label{equ:19}
  v_k = \frac{\mbox{max}(|z_k| + \sigma \bm{A}_k\bm{v}- \lambda,0)}{\eta +\sigma d_k}, k = 1,2,\cdots
\end{equation}
Let $\check{v}_k = \mbox{max}(|z_k| + \sigma \bm{A}_k\bm{v}- \lambda,0)$, and $\check{\bm{v}}=(\check{v}_1,...,\check{v}_p)^T$. To meet the normalizing condition, we let
$\tilde{\bm{v}}=\frac{\check{\bm{v}}}{\| \check{\bm{v}} \|_2}.$
Finally, based on Theorem 1, we get the optimal solution of Eq. (\ref{equ:12}) as follows:
\begin{equation}\label{equ:20}
  \bm{v}=\tilde{\bm{v}} \bullet \mbox{sign}(\bm{z}),
\end{equation}
where $``\bullet"$ denotes element-wise product. Likewise, fixed $\bm{u}$ in Eq. (\ref{equ:sgSVD*}), we can obtain the coordinate update for $\bm{u}$.
\begin{algorithm}[h]
\caption{\textbf{$L_1$-sgSVD*}.} \label{alg:Framwork1}
\begin{algorithmic}[1]
\REQUIRE Data matrix $\bm{X}\in \mathbb{R}^{n\times p}$; Prior networks $\bm{A^1}\in \mathbb{R}^{n\times n}$ and $\bm{A^2}\in \mathbb{R}^{p\times p}$; Parameters $\lambda_u, \lambda_v, \sigma_u, \sigma_v$.
\ENSURE $\bm{u}$, $\bm{v}$, $d$.
\STATE Initialize $\bm{v}$ with $\|\bm{v}\|_2=1$
\REPEAT
\STATE Let $\bm{z}=\bm{Xv}$, $\bm{A} = \bm{A^1}$ and $\bm{u} = |\bm{u}|$
\FOR {$i = 1$~to~$n$}
\STATE $u_i = \mbox{max}(|z_i| + \sigma_u \bm{A}_i\bm{u}- \lambda_u,0)$
\ENDFOR
\STATE $\bm{u}=\frac{\bm{u}}{\| \bm{u} \|_2}$
\STATE $\bm{u}=\bm{u} \bullet \mbox{sign}(\bm{z})$
\STATE Let $\bm{z}=\bm{X}^T\bm{u}$, $\bm{A} = \bm{A^2}$ and $\bm{v} = |\bm{v}|$
\FOR {$k = 1$~to~$p$}
\STATE $v_k = \mbox{max}(|z_k| + \sigma_v \bm{A}_k\bm{v}- \lambda_v,0)$
\ENDFOR
\STATE $\bm{v}=\frac{\bm{v}}{\| \bm{v} \|_2}$
\STATE $\bm{v}=\bm{v} \bullet \mbox{sign}(\bm{z})$
\STATE $d=\bm{z}^T\bm{v}$
\UNTIL $d$ convergence
\RETURN $\bm{u}$, $\bm{v}$, $d$.
\end{algorithmic}
\end{algorithm}

However, there are still two problems remain unresolved: (1) How to solve the singular value $d$? (2) What is the termination condition to stop the iterations? Note that the objective function of Eq.~(\ref{equ:sgSVD*}), $\|\bm{X} - d\bm{uv}^T\|_F^2 = \|\bm{X}\|_F^2-2d \bm{u}^T\bm{X}\bm{v}+d^2$. Given $\bm{u}$ and $\bm{v}$, this is a quadratic function about $d$. Hence the optimal solution $d^* = \bm{u}^T\bm{Xv}$, and the corresponding optimal value is $\|\bm{X}\|_F^2 - {d^*}^2$. Note that the minimum is only related with $d$. Therefore, we can control iteration by monitoring the change of $d$. Once the change of $d$ is smaller than a small positive constant $\epsilon$, where $\epsilon$ represents a tolerance level, we stop the iterations. Finally, we solve the optimization problem Eq. (\ref{equ:sgSVD*}) (Lagrangian form) by using an efficient Alternating Iterative Projection Algorithm (Algorithm 1). Note that the time complexity of Algorithm \ref{alg:Framwork1} is $\mathcal{O}(Tnp + Tp^2 + Tn^2)$, where $T$ is the number of iterations.

\subsection{$L_0$-norm Sparse Graph-regularized SVD}
In this section, we consider a novel sparse graph-regularized SVD via $L_0$-norm (denoted as $L_0$-sgSVD* or sgSVD*) as follows:
\begin{equation}
\begin{aligned}\label{equ:27}
& \underset{\bm{u},\bm{v},d}{\text{minimize}} && \|\bm{X} - d\bm{uv}^T\|_F^2\\
& \text{subject to}                         && \|\bm{u}\|_2 \leq 1, \|\bm{u}\|_0 \leq k_u, |\bm{u}|^T\bm{L_1}|\bm{u}| \leq c_1,\\
&                                           && \|\bm{v}\|_2 \leq 1, \|\bm{v}\|_0 \leq k_v, |\bm{v}|^T\bm{L_2}|\bm{v}| \leq c_2.
\end{aligned}
\end{equation}

To solve problem Eq. (\ref{equ:27}), we first fix $\bm{u}$ and let $\bm{z} = \bm{X}^T\bm{u}$. We then get the following optimization problem:
\begin{equation}
\begin{aligned}\label{equ:29}
& \underset{\bm{v}}{\text{minimize}} && -\bm{v}^T\bm{z}\\
& \text{subject to}                         && \|\bm{v}\|_2 \leq 1,\\
&                                           && \|\bm{v}\|_0 \leq k_v,\\
&                                           && |\bm{v}|^T\bm{L}|\bm{v}|\leq c_2.
\end{aligned}
\end{equation}
Similar with Theorem 1, we also have Theorem 3 to cleverly remove the absolute operator in the last constrained condition of Eq. (\ref{equ:29}).
\begin{theorem} Suppose $\bm{v}^*$ is an optimal solution of Eq.~(\ref{equ:29}), then $v^*_i z_i \geq 0$ for all $i$. \end{theorem}
Therefore, to solve Eq. (\ref{equ:29}), we first solve the following optimization problem,
\begin{equation}
\begin{aligned}\label{equ:29.5}
& \underset{\bm{v}}{\text{minimize}} && -\bm{v}^T|\bm{z}|\\
& \text{subject to}                  && \|\bm{v}\|_2 \leq 1,\\
&                                    && \|\bm{v}\|_0 \leq k_v,\\
&                                    && \bm{v}^T\bm{L}\bm{v} \leq c_2,\\
&                                    && v_k \geq 0, \forall k.
\end{aligned}
\end{equation}
We can further consider the following problem,
\begin{equation}
\begin{aligned}\label{equ:30}
& \underset{\bm{v}}{\text{minimize}} && f(\bm{v})\\
& \text{subject to}                  && \|\bm{v}\|_0 \leq k_v, 
\end{aligned}
\end{equation}
where $f(\bm{v}) = -\bm{v}^T|\bm{z}| + \frac{1}{2}\eta \bm{v}^T\bm{v} + \frac{1}{2} \sigma \bm{v}^T\bm{L}\bm{v}-\sum_{i=1}^p \tau_i v_i$, and $\eta \geq 0, \sigma \geq 0, \tau_i \geq 0$ are Lagrangian multipliers. Similar with Theorem 2, we can also prove $f(\bm{v})$ is a convex function. Note that when $\sigma = 0$, Eq. (\ref{equ:30}) corresponds to L0SVD. If we remove the constrained condition $\|\bm{v}\|_0 \leq k_v$ in problem Eq. (\ref{equ:30}), then we can obtain an update rule for $\bm{v}$ by using coordinate descent method \cite{friedman2007pathwise} (the derivation is similar to Eq. (\ref{equ:19})). The complementary slackness KKT condition implies that if $v_k > 0$, then $\tau_k = 0$ and if $v_k = 0$, then $\tau_k > 0$. Thus, we obtain the following update rule,
\begin{equation}\label{equ24}
  v_k = \frac{\mbox{max}(|z_k| + \sigma \bm{A}_k\bm{v},0)}{\eta +\sigma d_k}, k = 1,2,\cdots
\end{equation}
Let $\check{v}_k = \mbox{max}(|z_k| + \sigma \bm{A}_k\bm{v}, 0)$ and $\bm{v}=(\check{v}_1,...,\check{v}_p)^T$. To satisfy the condition $\|\bm{v}\|_0 \leq k_v$, we force the $p-k_v$ elements of $\bm{v}$ with the smallest absolute values to be zeros, i.e., project the vector $\bm{v}$ onto the closets absolute vector in Euclidean space,
\begin{equation}\label{equ25}
\bm{\widehat{v}} = \bm{v} \bullet I(|\bm{v}| \geq |\bm{v}|_{(k_v)}),
\end{equation}
where $I(\cdot)$ is the indicator function, $``\bullet"$ denotes element-wise product and $|\bm{v}|_{(k_v)}$ denotes the $k_v$-th order statistic of $|\bm{v}|$. To meet the normalizing condition, let $\bm{\widehat{v}} = \bm{\widehat{v}}/\|\bm{\widehat{v}}\|_2$. Based on Theorem 3, we can get a feasible solution of Eq. (\ref{equ:29}) as follows,
\begin{equation}\label{equ26}
  \bm{v}=\widehat{\bm{v}} \bullet \mbox{sign}(\bm{z}).
\end{equation}
Hence, we obtain an approximate solution of Eq. (\ref{equ:29}) by using a sparse projection strategy. Note that the iterative way of Eq. (\ref{equ25}) is consistent with the iterative threshold algorithm \cite{she2009thresholding}. In the same manner, for a fixed $\bm{v}$, we can also obtain the coordinate update of $\bm{u}$.

Finally, we propose an Alternating Iterative Sparse Projection (AISP) algorithm (Algorithm 2) to solve the optimization problem (\ref{equ:27}) (Lagrangian form). When $\sigma_1 = \sigma_2 =0$, $L_0$-sgSVD* algorithm reduces to L0SVD \cite{min2015novel}. We can prove the Eq. ($\ref{equ26}$) is optimal solution for rank one L0SVD model by fixing $\bm{u}$. In addition, prior graph information is considered to be more important with large $\sigma_u$ ($\sigma_v$). Moreover, Algorithm \ref{alg:Framwork2} has the same time complexity with Algorithm \ref{alg:Framwork1}.
\begin{algorithm}[h]
\caption{ \textbf{$L_0$-sgSVD*}.} \label{alg:Framwork2}
\begin{algorithmic}[1]
\REQUIRE Data matrix $\bm{X}\in \mathbb{R}^{n\times p}$; Prior networks $\bm{A^1}\in \mathbb{R}^{n\times n}$ and $\bm{A^2}\in \mathbb{R}^{p\times p}$; Parameters $k_u, k_v, \sigma_u, \sigma_v$.
\ENSURE $\bm{u}$, $\bm{v}$, $d$.
\STATE Initialize $\bm{v}$ with $\|\bm{v}\|_2=1$
\REPEAT
\STATE Let $\bm{z}=\bm{Xv}$, $\bm{A} = \bm{A^1}$ and $\bm{u} = |\bm{u}|$
\FOR {$i = 1$~to~$n$}
\STATE $\check{u}_i = |z_i| + \sigma_u \bm{A}_i\bm{u}$
\ENDFOR
\STATE $\bm{u} = \bm{\check{u}} \bullet I(|\bm{\check{u}}| \geq |\bm{\check{u}}|_{(k_u)})$
\STATE $\bm{\widehat{u}}=\frac{\bm{u}}{\| \bm{u} \|_2}$
\STATE $\bm{u}=\widehat{\bm{u}} \bullet \mbox{sign}(\bm{z})$
\STATE Let $\bm{z}=\bm{X}^T\bm{u}$ and $\bm{A} = \bm{A^2}$
\FOR {$k = 1$~to~$p$}
\STATE $\check{v}_k = |z_k| + \sigma_v \bm{A}_k\bm{v}$
\ENDFOR
\STATE $\bm{v} = \bm{\check{v}} \bullet I(|\bm{\check{v}}| \geq |\bm{\check{v}}|_{(k_v)})$
\STATE $\bm{\widehat{v}}=\frac{\bm{v}}{\| \bm{v} \|_2}$
\STATE $\bm{v}=\widehat{\bm{v}} \bullet \mbox{sign}(\bm{z})$
\STATE $d=\bm{z}^T\bm{v}$
\UNTIL $d$ convergence
\RETURN $\bm{u}$, $\bm{v}$, $d$.
\end{algorithmic}
\end{algorithm}

\begin{figure*}[htbp]
  \centering
  \includegraphics[width=1.0\linewidth]{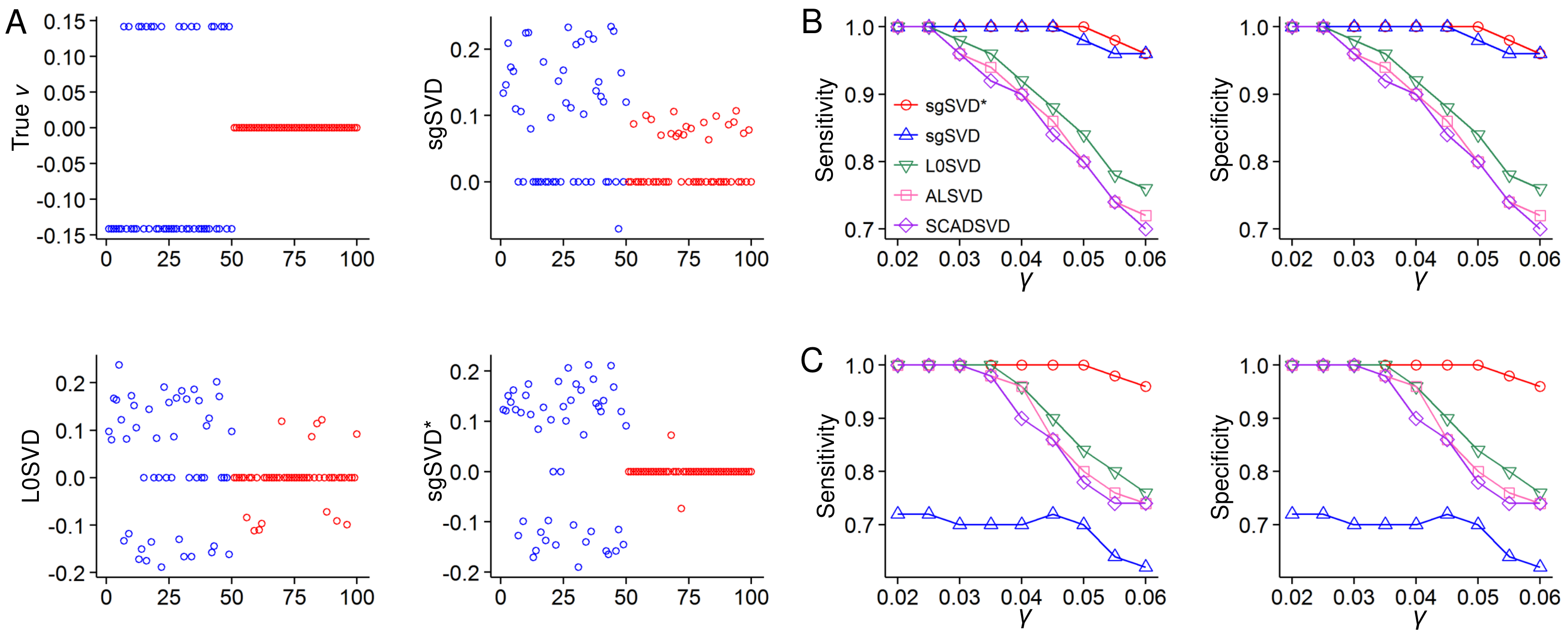}
  \caption{(A) Illustration of sgSVD*, sgSVD and L0SVD with a toy example ($\gamma = 0.06$) to explain the difference among them. (B, C) Evaluation of different methods using simulation data with two cases. (B) The signs of coefficients in left (or right) singular vector are forced to be the same. (C) The signs of coefficients in left (or right) singular vector are chosen at random. Note that we only show the results for $\bm{v}$ due to the symmetry of $\bm{u}$ and $\bm{v}$ in form.}\label{fig:toyExamples}
\end{figure*}
\section{Experiments}
\subsection{Synthetic Data Experiments}
Here, we evaluate the performance of sgSVD* with simulated data and compare its performance with sgSVD, L0SVD \cite{min2015novel}, ALSVD \cite{lee2010biclustering} and SCADSVD (sparse SVD via SCAD penalty) \cite{fan2001variable}.

Without loss of generality, we let a rank-one true signal matrix $\bm{X}^\ast = d \bm{u}\bm{v}^T$ and $d = 1$, where $\bm{u}$ and $\bm{v}$ are vectors of size $n \times 1$ and $p \times 1$, respectively. The observed matrix $\bm{X}$ is generated as the sum of $\bm{X}^\ast$ and a noise matrix $\bm{\epsilon}$, i.e., $\bm{X} = \bm{u}\bm{v}^T + \gamma \bm{\epsilon}$, where the elements of $\bm{\epsilon}$ are randomly sampled from a standard normal distribution (i.e., $\bm{\epsilon}_{ij}\sim \mathcal{N}(0,1)$), and $\gamma$ is a nonnegative parameter to control the signal-to-noise ratio. Accordingly, we generate two sparse singular vectors $\bm{u}$ and $\bm{v}$ with $n=p=100$ as follows:
$$\widetilde{\bm{u}} = [sample(\{-1,1\},50), rep(0,50)], \bm{u}=\widetilde{\bm{u}}/\|\widetilde{\bm{u}}\|_2, $$
$$\widetilde{\bm{v}} = [sample(\{-1,1\},50), rep(0,50)], \bm{v}=\widetilde{\bm{v}}/\|\widetilde{\bm{v}}\|_2, $$
where $sample(\{-1,1\},a)$ denotes a vector of size $a$, whose elements are sampled from $\{-1,1\}$, and $rep(0,b)$ denotes a vector of size $b$, whose elements are zeros. Here we create a data matrix $\bm{X}$ for each $\gamma$ ranging from 0.02 to 0.06 in steps of 0.005 and two prior graphs for row and column variables of $\bm{X}$, whose first 50 vertexes are connected with probability $p_{11}=0.3$, and remaining ones are connected with probability $p_{12}=0.1$. Moreover, to test sgSVD whether depends on the signs of singular vectors, we also simulate the data generated from the variables with the same signs by setting $\bm{u} := |\bm{u}|$ and $\bm{v}:=-|\bm{v}|$ with the same parameters (e.g., $n$, $p$, $\gamma$).

We apply the five methods to the simulation data with $\sigma_u = 0.1$ and $\sigma_v= 0.1$ which are selected by using cross validation strategy. We enforce each singular vector ($\bm{u}$ or $\bm{v}$) to contain 50 non-zeros elements (the same sparsity level) for each method by tuning the parameters to get comparable results for different methods. We first demonstrate the magnitude of singular vectors with absolute operator is how to overcome the opposite effect of classical graph-regularized penalty with a simple example which corresponds to Figure 1C with $\gamma = 0.06$ (Figure 1A). Moreover, we can clearly find that the performance on both sensitivity and specificity of sgSVD* and sgSVD are superior to that of L0SVD, ALSVD and SCADSVD (Figure~\ref{fig:toyExamples}B). As to the signals with diverse signs, sgSVD* demonstrates distinct better performance than sgSVD and others (Figure ~\ref{fig:toyExamples}C). These results suggest that (1) sgSVD* indeed show effectiveness compared to other methods (e.g., sgSVD and L0SVD); (2) sgSVD does suffer distinct limitations when some connected variables in the prior graph have different signal signs.
\begin{figure*} [htbp]
  \centering
  \includegraphics[width=1\linewidth]{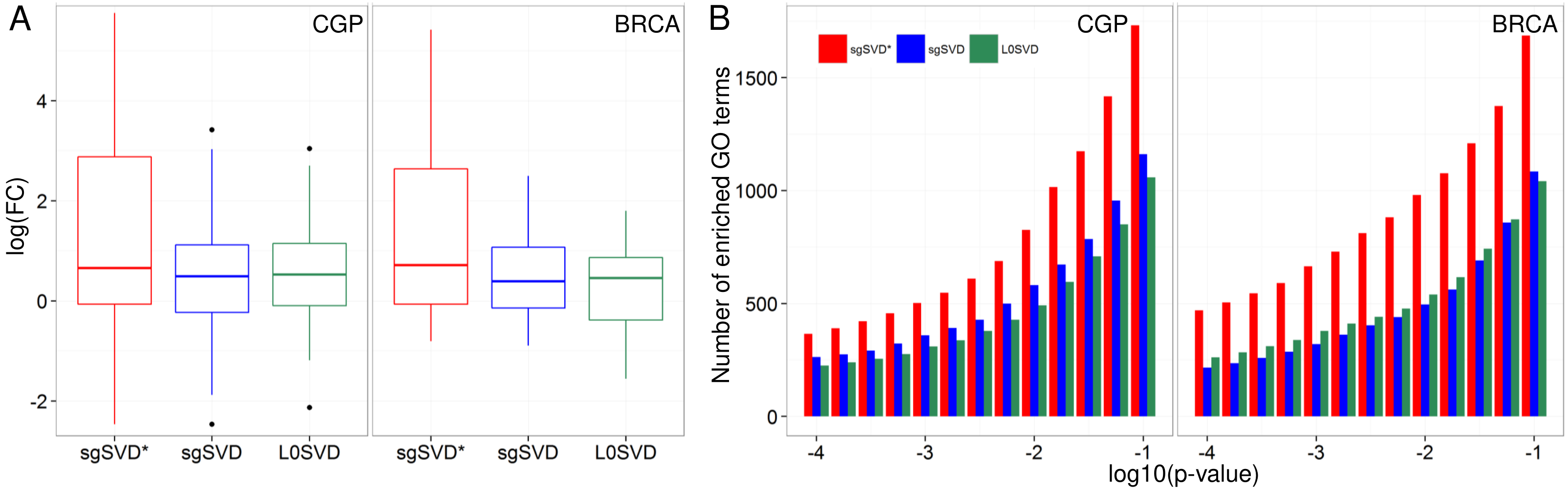}
  \caption{Comparison of three methods using CGP and BRCA data. (A) The distributions of log(FC) scores of identified modules by each method. sgSVD* shows distinct high enriched scores compared to those of sgSVD and L0SVD (Wilcoxon signed rank test $p$-value $<$ 0.01) on both data. (B) The number of significantly enriched functional terms of modules by each method with respect to different significant levels with log10($p$-value) from -4 to -1. sgSVD* shows distinct high enriched number of terms compared to those of sgSVD and L0SVD ($p$-value $<$ 0.01) on both data.  }\label{fig:comparison}
\end{figure*}
\subsection{Applications to Gene Expression Data}
We apply sgSVD* to two cancer gene expression datasets to demonstrate its effectiveness in deciphering sample-specific gene modules. We download a gene expression dataset from the Cancer Genome Project (CGP) \cite{garnett2012systematic} and obtain a data matrix with 13321 genes across 641 cell lines (samples). We further download a breast cancer gene expression dataset (BRCA) from TCGA database~\cite{cancer2012comprehensive}, and obtain a data matrix consisting of 11221 genes and 518 samples. For both applications, we construct a prior gene graph consisting of 13,321 genes and 262,462 interactions (a gene interaction network) from the Pathway-Commons database.

We apply sgSVD* to CGP and BRCA data with $k_u=200$ (genes), $\sigma_u = 0.4$ and a prior gene graph denoted as $\bm{A^1}$ for gene variables, $k_v=50$ (samples), $\sigma_v=0$, i.e., there is no prior graph for sample variables here. For each data, we first obtain the first pair singular vectors $\bm{u}$ and $\bm{v}$. Next, we subtract the signal of current pair singular vectors from the input data (i.e., $\bm{X}:=\bm{X}-d\bm{u}^T\bm{v}$), and apply sgSVD* again to identify the next pair singular vectors. The first 40 pair singular vectors are considered to identify 40 gene modules. Similarly, we also apply sgSVD and L0SVD to the CGP and BRCA data for comparison. We find that the gene pairs within each identified module do show distinct expression correlations by comparing the sum of absolute Pearson correlation coefficient between each pair within a module to those randomly generated ones. We also test $\sigma_v$ with different values and find the corresponding results are basically consistent. Herein below, we thus only show the results obtained with $\sigma_v=0.4$.

For a given module $i$ of $n_i$ genes and $m_i$ interactions (edge) in the prior graph, we assess the edge-enrichment of the module using a Fold Change (FC) score computed as FC = $\frac{m_i/\binom{n_i}{2}}{M/\binom{N}{2}}$, where $M$ is the total number of interactions and $N$ is the total number of genes in the prior gene graph. We can clearly see that that the FC scores of the modules identified from CGP and BRCA data by sgSVD* are significantly higher than those of sgSVD and L0SVD (Figure \ref{fig:comparison}A). In addition, the statistical significance of the FC is calculated by using the right tailed hypergeometric test. We find the percentages of edge-enriched gene modules of sgSVD* are much higher than that of sgSVD and L0SVD for different significance levels on both BRCA and CGP data (Table 1). These results suggest the modules identified using sgSVD* show distinct biological relevance than those of other methods.
\begin{table}[ht]
\centering
\caption{The percentage (\%) of significantly edge-enriched modules at different significance levels.}
\resizebox{\columnwidth}{!}{
\begin{tabular}{|ccc||ccc|c|}
   \hline
   &\textbf{BRCA}&&&\textbf{CGP}&&\\
   \hline
    L0SVD&sgSVD& sgSVD*& L0SVD& sgSVD& sgSVD*&Level\\
   \hline
    57.5&60.0& \textbf{70.0}& 57.5& 62.5&\textbf{67.5}& 0.10\\
   \hline
    57.5&57.5& \textbf{70.0}& 57.5& 60.0&\textbf{65.0}& 0.05\\
   \hline
    55.0&57.5& \textbf{65.0}& 57.5& 57.5&\textbf{62.5}& 0.01\\
   \hline
    55.0&57.5& \textbf{65.0}& 57.5& 57.5&\textbf{62.5}& 0.005\\
   \hline
    55.0&57.5& \textbf{65.0}& 55.0& 57.5&\textbf{60.0}& 0.001\\
   \hline
\end{tabular}\label{tab1}
}
\end{table}

To further valid our findings with biological functions, we also evaluate the functional enrichment of each module. We find that the gene modules identified by sgSVD* are significantly enriched in known Gene Ontology (GO) biological process (BP) terms at different significance levels than those of sgSVD and L0SVD in both data (Figure \ref{fig:comparison}B). In addition, we find that the number of enriched modules for most significance levels of sgSVD are lower than those of L0SVD in BRCA data. It is probably because the coefficients of a number of variables linked on the prior graph have different signs in this data. Furthermore, to assess whether the number of significant GO biological processes are in favor of chance, we also perform the same test on forty random modules whose gene labels are randomly permuted. However, no significant GO BP term is discovered. All these results demonstrate that sgSVD* can identify more biologically relevant gene modules than other methods.

\section{Conclusion}
In this paper, we propose a $L_0$-norm sparse graph-regularized SVD for biclustering high-dimensional data. We adopt the graph-regularized penalty and $L_0$-norm to induce structure sparsity of singular vectors and enhance intelligibility of data structure. Importantly, we consider the magnitude of singular vectors with absolute operator to overcome the opposite effect of classical graph-regularized penalty. To solve this novel model, we first find a clever trick to remove the absolute operator in the new graph-regularized penalty, and then design an efficient Alternating Iterative Sparse Projection (AISP) algorithm to solve it. We expect the proposed method can be applied to high-dimensional data from diverse domains. Moreover, it will be valuable to extend current concept into other statistical learning frameworks [e.g., Canonical Correlation Analysis (CCA), partial least square (PLS)].

\section*{Acknowledgment}
This work was supported by the National Science Foundation of China [61379092, 61422309], Natural Science Foundation of Hubei Province of China 2014CFB194, and the Outstanding Young Scientist Program of CAS, and the Key Laboratory of Random Complex Structures and Data Science, CAS.

\bibliographystyle{named}
\balance
\bibliography{ijcai16}
\end{document}